\pdfoutput=1
\documentclass[11pt]{article}

\usepackage[final]{acl}

\usepackage{times}
\usepackage{enumitem}
\usepackage{latexsym}
\usepackage{amsmath}
\usepackage{array}

\usepackage[T1]{fontenc}
\usepackage[utf8]{inputenc}

\usepackage{microtype}

\usepackage{inconsolata}

\usepackage{graphicx}

\title{FB-RAG: Improving RAG with Forward and Backward Lookup}

\author{Kushal Chawla, Alfy Samuel, Anoop Kumar, Daben Liu\\
  Capital One \\
  \texttt{\{kushal.chawla,alfy.samuel,anoop.kumar,daben.liu\}@capitalone.com}}

\begin{document}
\maketitle
\begin{abstract}

% Retrieval Augmented Generation (RAG) allows Large Language Models (LLMs) to ground their responses on domain-specific external knowledge, which has shown promise in reducing hallucinations and improving performance across diverse tasks and application domains.

Traditional Retrieval-Augmented Generation (RAG) struggles with complex queries that lack strong signals to retrieve the most relevant context, forcing a trade-off between choosing a small context that misses key information and a large context that confuses the LLM. To address this, we propose \textbf{Forward-Backward RAG (FB-RAG)}, a new training-free framework based on a simple yet powerful \textit{forward-looking} strategy. FB-RAG employs a light-weight LLM to \textit{peek} into potential future generations, using evidence from multiple sampled outputs to precisely identify the most relevant context for a final, more powerful generator. This improves performance without complex finetuning or Reinforcement Learning common in prior work. Across $9$ datasets from LongBench and $\infty$Bench, FB-RAG consistently delivers strong results. Further, the performance gains can be achieved with reduced latency due to a shorter, more focused prompt for the powerful generator. On \textbf{EN.QA} dataset, FB-RAG matches the leading baseline with over $48$\% latency reduction or achieves an $8$\% performance improvement with a $10$\% latency reduction. Our analysis finds cases where even when the forward-looking LLM fails to generate correct answers, its attempts are sufficient to guide the final model to an accurate response, demonstrating how smaller LLMs can systematically improve the performance and efficiency of larger ones. Our code is available at: \url{https://github.com/CapitalOne-Research/fb-rag}.

\end{abstract}

\section{Introduction}
Retrieval-Augmented Generation (RAG) shows immense promise in reducing hallucinations and improving generation performance~\cite{fan2024survey,gao2023retrieval}. RAG achieves strong results on diverse Question Answering (QA) tasks~\cite{borgeaud2022improving,guu2020retrieval,asai2024self}, general language tasks~\cite{he2021fast,khandelwalgeneralization}, and across numerous downstream applications~\cite{liu2023multi,wu2024coral}.

\begin{figure*}[t]
\centering
  \includegraphics[width=\linewidth]{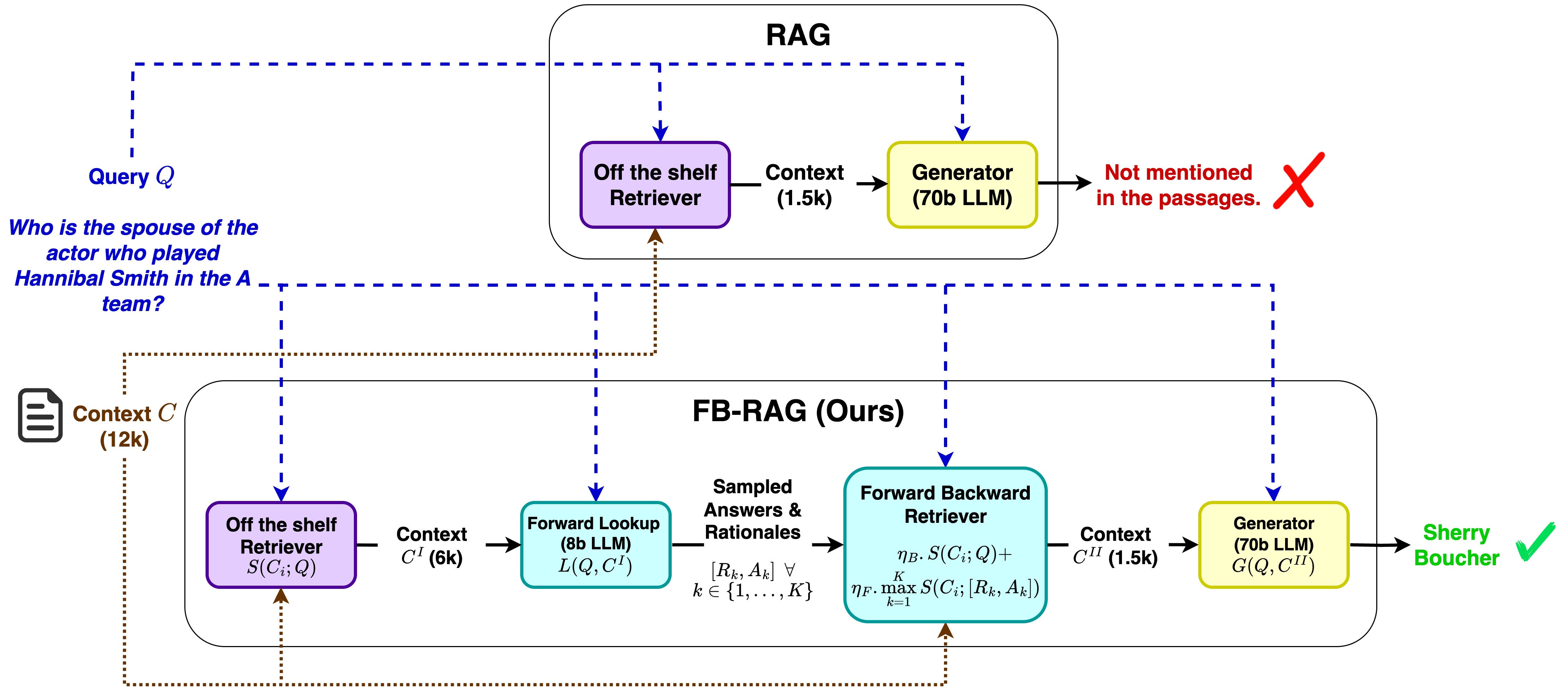}
  \caption{Overview of \textbf{FB-RAG}: a training-free framework for generating answers for an input query and context. FB-RAG looks at both the input query and sampled outputs from a light-weight LLM to rank context chunks.}
  \label{fig:fbrag-overview}
\end{figure*}

In this work, we focus on the task of answering queries based on an \textit{already-provided} large context. Traditional RAG efforts for this setup involve two steps~\cite{zhao2024retrieval}: 1) Retrieve important chunks by computing similarities with the query (based on a sparse or dense retriever and/or a reranker), 2) Feed the retrieved chunks along with the query to an LLM, which generates the answer. We refer to these approaches as \textit{backward-looking} -- looking back at the input query to score the context chunks. Such methods have been widely adopted in both academia and industry. However, standard methods struggle with complex queries that lack sufficient information to retrieve relevant chunks (see example in Figure ~\ref{fig:fbrag-overview}). This challenge is difficult to manage in RAG, where retrieving too little risks missing key information and retrieving too much risks adding irrelevant content that can confuse the LLMs~\cite{yu2024defense}.

%%This challenge is challenging  to manage with the duality that naturally arises in RAG - retrieving too little risks missing key information and retrieving too much risks adding irrelevant content that can confuse the LLMs~\cite{yu2024defense}.

To address this challenge, we design \textbf{Forward-Backward RAG} (\textbf{FB-RAG}) for studying the impact of an emerging yet underexplored idea -- \textit{forward-looking} or \textit{peeking} into the LLM's output generations to improve retrieval. \textbf{FB-RAG} generates the output in three stages: \textbf{I)} \textit{Recall-focused Retrieval}, using an off-the-shelf retriever to extract a smaller, yet sufficiently large context, \textbf{II)} \textit{Precision-focused Retrieval}, which either only relies on \textit{forward-looking} by observing reasons and answers from a light-weight LLM to evaluate the context chunks (Ours-F) or relies on both \textit{forward} and \textit{backward} lookup (Ours-FB), and \textbf{III)} \textit{Generation}, prompting a powerful LLM to get the final answer. Although prior work used related ideas to improve RAG with LLM-based feedback or confidence scores~\cite{zhao2024longrag, sun2022recitation, wang2024speculative, yang2023prca, jiang-etal-2023-active}, these methods typically propose complex fine-tuning or Reinforcement Learning strategies, and often assume access to external web search engines or rely on LLM's own memory which is not suitable for many domain-specific practical settings. Instead, \textbf{FB-RAG} is a simple and effective training-free framework based on off-the-shelf retrievers and instruction-tuned LLMs that answers questions from an already-provided large context.

To find relevant chunks with an \textit{imperfect} forward-looking LLM, \textbf{FB-RAG} samples multiple outputs and assigns a high score to a chunk if it was used for \textit{any} of them. This turns out to be powerful for improving RAG results over recent baselines across diverse tasks. We also find that gains can be achieved while reducing latency. On \textbf{EN.QA} dataset from $\infty$Bench~\cite{zhang-etal-2024-bench}, one can combine a 70B parameter model for final response generation with an 8B model for forward-lookup and match the baseline performance with over $48$\% latency reduction. Further, one can get an $8$\% performance improvement with $10$\% latency reduction. Through our qualitative analysis, we find instances where even if all the sampled outputs from the smaller LLM incorrectly answer the input query, and often fail to follow our instructions properly, this still proves sufficient for the final, more powerful LLM to generate the correct response. We now summarize our contributions:
\begin{enumerate}
    \item We propose \textbf{FB-RAG}: a training-free framework for performing RAG with off-the-shelf instruction-tuned LLMs. FB-RAG employs a simple and effective look-ahead strategy to evaluate context chunks before selecting them for final response generation (Section \ref{sec:methodology}).
    \item We comprehensively evaluate FB-RAG against recent training-free RAG and Long Context baselines on $9$ datasets from LongBench~\cite{bai-etal-2024-longbench} and $\infty$Bench~\cite{zhang-etal-2024-bench}, finding that FB-RAG delivers consistent performance improvements. We further analyze key design choices in FB-RAG, such as the number of chunks retrieved and the number of samples used for forward lookup (Sections \ref{sec:expt-design} and \ref{sec:results}).
    \item We show that FB-RAG provides the flexibility to improve performance while reducing latency. We additionally perform qualitative analysis discussing the strengths and limitations of our approach, and provide insights for future progress in this area (Section \ref{sec:discussion}).
\end{enumerate}

% recent efforts have increased long context lengths....and related benchmarks to test the models capability like longbench, infinity bench.

% Prior work has observed scenarios where Long context outperforms RAG and vice versa. Observed a U shape on two datasets from Inf Bench -> small context lengths ...and long more irrelevant information.

% one thing is clear --> it is generally better to have the right answer in a smaller context..If we could come up with a more precise context, that can lead to improved performance, surpassing long context baselines. backward-looking fails.

% To this end, we propose Forward-Backward RAG --> . define forward and backward a three step approach Step 1 Recall focused ->to bring down the context..
% 2 is precision focused that looks at both We now summarize our contributions..acts as a retriever.

% Say we evaluate on 9 datasets and show performance improvements over long context and other rag baselines. We present analysis on how the prompt ..etc impacts..

% List the contributions and add section numbers.

% Long context length is clear --> A competing paradigm is RAG --> 

% not robust to irrelevant context. talk about U shape..better to have it in there in a small context...talk about little information in the query which fails.. self route has good language.

% What is the story here?v talk about long context that token lengths have increased.

\section{Methodology}
\label{sec:methodology}

We focus on the task of question answering based on an \textit{already-provided} context. Given an input query $Q$ and a context $C$, FB-RAG relies on an off-the-shelf retriever and instruction-tuned LLMs (without finetuning) to generate the output $M(Q, C)$\footnote{This general formulation encompasses several QA, summarization, and Multiple Choice Questions (MCQ) tasks - see Section~\ref{sec:expt-design} for the datasets considered in this work.}. We assume that context $C$ is sufficient to answer the query $Q$, differentiating from some prior formulations that assume runtime access to web search engines~\cite{yan2024corrective}. At its core, FB-RAG relies on a look-ahead method to retrieve the most relevant context chunks from $C$ before performing the final response generation. We start by describing this method and later connect it to the overall three-stage process of FB-RAG.
\subsection{Forward-Backward Retriever}
\label{sec:fb-retriever}
We are given a query $Q$ and context $C = \{C_i\} = \{C_1, C_2, C_3, ... C_{n}\}$, with $n$ chunks in total. We use $A^*$ to denote the ideal output response (ground-truth answer), and $C_i^* \in C$ to denote the context chunk that contains the information needed to generate the ideal answer $A^*$. Further, we use $S(C_i ; Q)$ to represent the importance score of a context chunk $C_i$ given a query $Q$ using an off-the-shelf retriever $S$. We use $S_{FB}(C_i ; Q,C)$ to denote the importance score of chunk $C_i$ under FB-RAG given a query $Q$ and the full associated context $C$. As in a typical RAG pipeline, once the importance scores are computed, we can select the highest-scoring chunks for final output generation using an LLM. Hence, our goal in this section is simply to provide a formulation for $S_{FB}(C_i ; Q,C)$.

Prior work has reported that LLMs often get confused by the irrelevant information present in the context~\cite{xu2024recomp,asai2024self}. The inverted U shape for the performance observed by~\citet{yu2024defense} as the context size increases demonstrates this in action. Hence, one obvious objective for the retrievers is to assign high importance scores to the most relevant chunks so that one can use a small context for generation and reduce irrelevant content. This is challenging for retrievers relying solely on the information in the input query, especially when the query is non-specific and complex~\cite{li-etal-2024-retrieval}. To address this gap, our key idea is to look \textit{forward} at the potential answer to retrieve the relevant contexts. If we had access to the oracle generation model $L^*$, we could compute $S_{FB}(C_i;Q,C)$ in the following manner:
\begin{equation}
    S_{FB}(C_i;Q,C) = S(C_i; L^*(Q, C)) = S(C_i;A^*).
\end{equation}
Unfortunately, even though we are using the oracle generator $L^*$, this formulation is still not sufficient. Oftentimes in QA, the answers are concise entities or even binary (\textit{yes} or \textit{no}), meaning that even the ideal answer $A^*$ might be insufficient to identify the most relevant context chunk $C_i^*$. Hence, we also enable the oracle to generate the ideal \textit{reasoning} $R^*$ before generating the final answer $A^*$:
\begin{align}
    S_{FB}(C_i;Q,C) &= S(C_i;L^*(Q, C)) \nonumber \\
                &= S(C_i;[R^*, A^*]).
\end{align}
For a reasonable retriever $S$, we now hypothesize:
\begin{equation}
    \arg\max_{i} S(C_i;[R^*, A^*]) = C_i^*,
\end{equation}
meaning that one can reasonably expect to reach $C_i^*$ if given access to the ideal reasoning $R^*$ and ideal answer $A^*$. Note that our assumption that there is a single chunk $C_i^*$ which contains all the relevant information to generate $A^*$ is not limiting; one can trivially extend the same argument to the case where the relevant information is split across multiple chunks. In such a case, we reasonably expect the most relevant chunks to be ranked higher than irrelevant chunks based on $S(C_i;[R^*, A^*])$.

We now approximate the oracle $L^*$ with an instruction-tuned LLM L:
\begin{align}
    S_{FB}(C_i;Q,C) &= S(C_i;L(Q, C)) \nonumber \\
                &= S(C_i;[R, A]),
\end{align}
where $R$ and $A$ are the reasoning and answer generated by the LLM $L$. To capture the uncertainty of the \textit{imperfect} LLM $L$, we further propose to consider the maximum over $K$ samples generated from the model:
\begin{equation}
\label{eq:fb-component}
    S_{FB}(C_i;Q, C) = \max_{k=1}^KS(C_i;[R_k, A_k]),
\end{equation}
where $R_k$ and $A_k$ are reasoning and answer in the $k^{th}$ sample respectively. Taking the maximum ensures that even if a chunk $C_i$ is used only in one sample, it will still receive a high score under $S_{FB}(C_i;Q, C)$. This is useful to capture the relevant chunks in cases where the LLM $L$ is not confident, resulting in high variance in the samples.

Equation \ref{eq:fb-component} utilizes the complete \textit{forward-looking} component of our framework (referred as $S_F$ below). However, our formulation of $S_{FB}(C_i ; Q,C)$ is still incomplete. In case of an extremely noisy model $L$, the generated reasoning sequences and corresponding answers can be inaccurate and thus, can provide a misleading signal for our purpose of ranking the context chunks. Hence, merely relying on the outputs from such a noisy model $L$ can unfairly penalize the true relevant chunk $C_i^*$. Motivated by this, we also incorporate a \textit{backward-looking} component (as a form of a regularizer) that looks at the original input query $Q$ to compute the importance scores:
\begin{align}
\label{eq:fb-weighted}
    &S_{FB}(C_i;Q, C) = \eta_B.S_B + \eta_F.S_F = \nonumber \\
    &\eta_B.S(C_i;Q) + \eta_F.\max_{k=1}^KS(C_i;[R_k, A_k]),
\end{align}
where $S_B$ and $S_F$ denote the \textit{backward} and \textit{forward} components respectively, while $\eta_B$ and $\eta_F$ refer to their corresponding weights. Equation \ref{eq:fb-weighted} completes our goal for this Section.

The \textit{forward} component $S_F$ relies on (reasoning, answer) samples generated by the LLM, which can be time-consuming as is. Of course, one can generate these samples in parallel, but we propose two additional simple solutions to manage this cost. \textbf{First,} the LLM used for this look-ahead can be selected independently from the LLM that is used to perform the final generation. In our experiments presented in Section~\ref{sec:results}, we use a \textit{relatively} light-weight LLM ($8$B parameters) for forward-lookup and a much more powerful LLM ($70$B parameters) for the final response generation. We also present results with other light-weight LLM choices later in Section~\ref{sec:discussion}. \textbf{Second,} one can use a fast retriever to reduce the context size before utilizing the Forward-Backward procedure laid out in this Section. These remedies motivate the three-step process of FB-RAG, which we describe below.

\subsection{FB-RAG Overview}
\label{sec:approach-overview}
We present our approach in Figure \ref{fig:fbrag-overview}. FB-RAG follows a three-stage process to compute the output response $M(Q,C)$: 1) \textit{Recall-focused Retrieval}, 2) \textit{Precision-Focused Retrieval}, and 3) \textit{Generation}.

\noindent\textbf{Recall-focused Retrieval}: In Stage I, we employ an off-the-shelf retriever to reduce the context size from $C$ to $C^{I}$. This is \textit{recall-focused}, meaning one can select a relatively large context while still reducing the size significantly compared to $C$. The goal here is not to perform generation with $C^{I}$, but rather to use it for Stage II.

\noindent\textbf{Precision-Focused Retrieval}: In Stage II, we follow the procedure laid out in Section~\ref{sec:fb-retriever} using a light-weight LLM $L$ to compute $S_{FB}(C_i;Q, C^{I})$. Importantly, $C_i$ still comes from the full input context $C$. We use these scores to \textit{precisely} select the relevant context chunks, reducing $C$ to $C^{II}$, which is our target context to be used for generation.

\noindent\textbf{Generation}: Lastly, we prompt another instruction-tuned LLM $G(Q, C^{II})$ to generate the final answer.

We evaluate two variants of \textbf{FB-RAG} in this paper: 1) \textbf{Ours-FB}: Using both $\eta_B$ and $\eta_F$ as $0.5$ in Equation \ref{eq:fb-weighted}, and 2) \textbf{Ours-F}: Using $\eta_B=0$ and $\eta_F=1$ (ignoring the backward-component and resorting to the formulation in Equation \ref{eq:fb-component} instead). As presented later in Sections \ref{sec:results} and \ref{sec:discussion}, we find that \textbf{Ours-F} consistently outperforms \textbf{Ours-FB} across the board, indicating that one needs to only rely on the forward-looking component -- at least for the choices for LLM $L$ considered in this work.

We make two observations about the overall performance achievable by our framework. \textbf{First}, the performance is not limited by $L(Q, C^{I})$ since the outputs from $L$ are only used \textit{softly} to score the chunks in the full context $C$, and the final generation is still performed by a more powerful LLM $G$. \textbf{Second}, the performance is also not limited by $G(Q, C^{I})$ since Stage II works to improve the position of $C_i^*$, \textit{increasing the likelihood} that $C_i^*$ is picked up in the \textit{smaller} context $C^{II}$, which can make it easier for $G$ to generate an accurate answer. We provide a deeper probabilistic interpretation of our approach in Appendix \ref{sec:appendix-methodology} and validate these observations empirically in Section~\ref{sec:results}.

\section{Experiment Design}
\label{sec:expt-design}
We address the following four research questions: \textit{\textbf{RQ 1)} Performance:} \textit{How does FB-RAG perform compared to RAG and Long Context baselines?} -- We evaluate FB-RAG on $9$ datasets spanning QA, Summarization, and Multiple Choice Questions (MCQ) tasks. \textit{\textbf{RQ 2)} Design Considerations:} \textit{What is the impact of key design choices on the performance of FB-RAG?} - We study the performance by varying the number of retrieved chunks, the number of samples used in Stage II, and the LLM used for forward lookup. \textit{\textbf{RQ 3)} Impact on Latency:} \textit{How does the three-stage process of FB-RAG impact the overall latency?} - We plot the performance against latency by varying the chunks and comparing our approach to a baseline. \textit{\textbf{RQ 4)} Qualitative Analysis:} \textit{In what specific scenarios does FB-RAG improve performance and what kind of errors does the approach make?} - We perform error analysis and discuss our insights for future work.

\noindent\textbf{Datasets:} Following prior work~\cite{li-etal-2024-retrieval}, we focus on tasks that are a) in English, b) real, and c) query-based. This leads to $7$ datasets from LongBench~\cite{bai-etal-2024-longbench}: \textbf{NarrativeQA}~\cite{kocisky-etal-2018-narrativeqa}, \textbf{Qasper}~\cite{dasigi-etal-2021-dataset}, \textbf{MultiFieldQA}~\cite{bai-etal-2024-longbench}, \textbf{HotpotQA}~\cite{yang-etal-2018-hotpotqa}, \textbf{2WikiMultihopQA}~\cite{ho2020constructing}, \textbf{MuSiQue}~\cite{trivedi2022musique}, and \textbf{QMSum}~\cite{zhong-etal-2021-qmsum}. We also pick two datasets from $\infty$Bench~\cite{zhang-etal-2024-bench}, namely, \textbf{En.QA} and \textbf{EN.MC}. These datasets cover diverse domains, including Wikipedia articles, meetings, narratives, and research papers, involving single and multi-hop questions. The average context lengths range from a few thousand to $150$k words. Refer to Appendix~\ref{sec:appendix-datasets} for more details.

\noindent\textbf{Metrics:} We use F1 score for QA datasets, Rouge-L F1 for summarization, and classification accuracy for the MCQ task. Our implementation is based on the code released with LongBench\footnote{\url{https://github.com/THUDM/LongBench/tree/main}}.

\noindent\textbf{Methods:} \textbf{Long Context (LC)} refers to directly feeding the full context to the LLM without explicit
retrieval. \textbf{Vanilla} denotes the typical RAG approach, which performs retrieval based on an off-the-shelf retriever before feeding the context to the LLM. We implemented two recent approaches evaluated on the considered datasets. In Order-Preserving (\textbf{OP}) RAG~\cite{yu2024defense}, the selected chunks from the retriever are first sorted in their original ordering before feeding them to the LLM. \textbf{Self-Route}~\cite{li-etal-2024-retrieval} is a look-ahead approach that relies on LLM's ability to understand if the question is answerable from the retrieved context. It involves $3$ steps: 1) \textit{Retrieval}: Based on an off-the-shelf retriever, 2) \textit{Generation:} A modified generation based on the retrieved context where the LLM can choose to output `unanswerable' if it finds that the retrieved context is insufficient to answer the question, and 3) \textit{Generation:} Based on the full input context if the LLM outputs `unanswerable' in the previous step.

For our approach, both \textbf{Ours-FB} and \textbf{Ours-F} variants use $5$ samples in Stage II obtained by combining top-p (p=$0.9$) and top-k (k=$50$) sampling. The final response generation for all methods uses Llama3.1-70B-Instruct~\cite{llama31-2024}. \textbf{Self-Route} uses the same model for both generation steps. For our approach, we use Llama3.1-8B-Instruct~\cite{llama31-2024} for generating samples in Stage II. Refer to Appendix~\ref{sec:appendix-expt-design} for the prompts used, hardware details, and token limits. We evaluated $4$ retrievers: BM25~\cite{trotman2014improvements}, M3Flag~\cite{chen-etal-2024-m3}, BGEFlag~\cite{xiao2024c}, and MPNet\footnote{\url{https://huggingface.co/sentence-transformers/multi-qa-mpnet-base-cos-v1}}. We chose BM25 for our experiments due to its strong relative performance, simplicity, and versatility, making it suitable for our approach, which relies on LLM-generated outputs to retrieve relevant context chunks (see Appendix \ref{sec:appendix-retriever-comparison} for a performance comparison). For chunking, we use a chunk size of $300$ words throughout.

\section{Results}
\label{sec:results}

\begin{table*}
  \centering
  \scalebox{0.8}{
  \begin{tabular}{l|c|ccccccc}
    \hline
    \textbf{Method} & \textbf{Avg} & \textbf{Narr} &  \textbf{Qasp} & \textbf{Mult} & \textbf{Hotp} & \textbf{2Wiki} & \textbf{Musi} & \textbf{QMSum} \\
    \hline
    \multicolumn{9}{c}{Long Context} \\
    % Gemini-1.5-Pro & 45.53 & 32.76 & 47.83 & 52.33 & 61.85 & 62.96 & 40.22 & 20.73\\
    % GPT-4O & 47.04 & 32.78 & 44.54 & 55.28 & 62.42 & 70.69 & 41.65 & 21.92\\
    Llama3.1-70B-Instruct & 49.28 & 33.42 & 50.96 & 55.63 & 64.4 & 67.18 & 48.68 & 24.68\\ \hline
    \multicolumn{9}{c}{Self-Route~\cite{li-etal-2024-retrieval}} \\
    Gemini-1.5-Pro & 43.33 & 28.32 & 45.23 & 51.47 & 55.18 & 62.68 & 40.66 & 19.77\\
    GPT-4O & 46.83 & 31.36 & 47.99 & 53.17 & 62.14 & \textbf{70.14} & 41.69 & 21.31\\ \hline
    \multicolumn{9}{c}{Llama3.1-70B-Instruct; RAG - Our Impl. (1.5k)} \\
     Vanilla & 44.19 & 25.01 & 49.31 & 53.41 & 60.91 & 58.84 & 37.32 & 24.51\\
    OP~\cite{yu2024defense} & 44.34 & 23.89 & 49.31 & 54.8 & 61.11 & 59.06 & 37.94 & 24.26\\
    Self-Route~\cite{li-etal-2024-retrieval} & 47.23 & 24.04 & 48.77 & 54.34 & 64.42 & 68.23 & 46.68 & 24.14\\
    Ours-FB (6k $\to$ 1.5k) & 49.36 & 30.29 & \textbf{51.38} & 56.22 & \textbf{68.76} & 63.27 & 50.92 & 24.68\\
    Ours-F (6k $\to$ 1.5k) & 49.36 & 28.62 & 51.29 & 55.53 & 66.99 & 65.1 & 52.93 & 25.07 \\ \hline
    \multicolumn{9}{c}{Llama3.1-70B-Instruct; RAG - Our Impl. (3k)} \\
    Vanilla & 47.09 & 26.99 & 50.55 & 54.67 & 65.33 & 61.06 & 46.55 & 24.48\\
    OP~\cite{yu2024defense} & 48.03 & 26.62 & 50.71 & 56.78 & 66.28 & 64.8 & 45.91 & 25.11\\
    Self-Route~\cite{li-etal-2024-retrieval} & 48.29 & 27.54 & 50.09 & 56.1 & 65.64 & 66.02 & 47.75 & 24.9\\
    Ours-FB (6k $\to$ 3k) & 50.23 & 33.22 & 50.99 & 55.99 & 66.29 & 67.42 & 53.13 & 24.56\\
    Ours-F (6k $\to$ 3k) & 50.31 & 32.41 & 51.05 & 56.12 & 66.79 & 67.95 & \textbf{53.7} & 24.17\\ \hline
    \multicolumn{9}{c}{Llama3.1-70B-Instruct; RAG - Our Impl. (6k)} \\
    Vanilla & 48.59 & 31.09 & 50.12 & 55.17 & 66.39 & 65.9 & 46.72 & 24.72\\
    OP~\cite{li-etal-2024-retrieval} & 48.75 & 29.85 & 51.35 & 55.6 & 65.53 & 65.5 & 48.85 & 24.59\\
    Self-Route~\cite{li-etal-2024-retrieval} & 48.71 & 30.52 & 50.74 & 54.67 & 66.5 & 64.12 & 49.29 & \textbf{25.13}\\
    Ours-FB (6k $\to$ 6k) & 50.05 & 33.24 & 50.87 & 56.57 & 65.25 & 67.76 & 51.94 & 24.75\\
    Ours-F (6k $\to$ 6k) & \textbf{50.51} & \textbf{34.36} & 50.84 & \textbf{57.26} & 65.36 & 67.63 & 53.4 & 24.69 \\ \hline
  \end{tabular}}
  \caption{Results on LongBench. We use Rouge-L F1 for \textbf{QMSum}, and F1 score for others. (X $\to$ Y): Context size $X$ in Stage I and $Y$ in Stage II. Comparisons with other popular retrievers are in Appendix \ref{sec:appendix-retriever-comparison}.}
  \label{tab:results-longbench}
\end{table*}

\begin{table}[t]
  \centering
  \scalebox{0.8}{
  \begin{tabular}{lcc}
    \hline
    \textbf{Method} & \textbf{EN.QA} & \textbf{EN.MC} \\
    \hline
    \multicolumn{3}{c}{Long Context} \\
    % Gemini-1.5-Pro & 43.08 & 85.57\\
    % GPT-4O & 32.36 & 78.42\\
    Llama3.1-70B-Instruct & 34.26 & 71.62\\ \hline
    \multicolumn{3}{c}{Self-Route~\cite{li-etal-2024-retrieval}} \\
    Gemini-1.5-Pro & 37.51 & 76.86\\
    GPT-4O & 34.95 & 77.29\\ \hline
    \multicolumn{3}{c}{Llama3.1-70B-Instruct; OP RAG~\cite{yu2024defense}} \\
    16k & 44.43 & 84.72\\
    24k & 45.45 & \textbf{88.65}\\
    48k & 47.25 & 85.59\\ \hline
    \multicolumn{3}{c}{Llama3.1-70B-Instruct; OP RAG (Our Impl.)} \\
    16k & 47.87 & 81.22\\
    24k & 48.27 & 85.59\\ \hline
    \multicolumn{3}{c}{Llama3.1-70B-Instruct; FB-RAG (Ours)} \\
    Ours-FB (24k $\to$ 12k) & 49.93 & 84.28\\
    Ours-FB (24k $\to$ 16k) & 51.68 & 85.59\\
    Ours-F (24k $\to$ 12k) & 50.38 & 85.59\\
    Ours-F (24k $\to$ 16k) & \textbf{52.24} & 86.46\\ \hline
  \end{tabular}}
  \caption{Results on $\infty$Bench. We report F1 score for \textbf{EN.QA} and accuracy for \textbf{EN.MC}. (X $\to$ Y): Context size $X$ in Stage I and $Y$ in Stage II.}
  \label{tab:results-infbench}
\end{table}

\noindent\textbf{FB-RAG outperforms Long Context and other RAG baselines on both LongBench and $\mathbf{\infty}$Bench datasets.} We present the main results on LongBench datasets in Table~\ref{tab:results-longbench}. Across diverse domains and context size settings, we find that our approach exhibits consistent performance improvements over other implemented methods. Our approach achieves the best score on $5$ out of $7$ datasets, and our method \textbf{Ours-F (6k $\to$ 6k)}, which uses a 6k context output in both Stage I and II, achieves the best average performance of $50.51$. We present the results for $\infty$Bench datasets in Table \ref{tab:results-infbench}. We find performance improvements on both datasets. Our approach \textbf{F (24k $\to$ 16k)} achieves $52.24$ on EN.QA outperforming both the top reported results in the OP RAG paper ($47.25$) and the best OP RAG result found in our own implementation ($48.27$). On EN.MC, our approach achieves $86.46$, which beats the best achieved in our implementation of OP-RAG ($85.59$) but does not beat the reported best result of $88.65$, potentially due to differences in the experiment design, such as the retriever and chunking methods.

%When the underlying LLM is less effective, the sampled reasoning and answers can be noisy and misleading. In these situations, we expect Ours-FB approach to perform better than Ours-F.

\noindent\textbf{Only looking forward in Stage II of FB-RAG generally performs better than averaging out Forward and Backward components.} We observe that setting $\eta_B=0$ in Equation \ref{eq:fb-weighted} (nullifying the backward-looking component in Stage II) performs better than giving equal weight to both forward and backward looking components. This indicates that when LLM-generated reasoning and answer samples are incorporated, the input query does not seem to provide any new useful information to retrieve the most relevant context chunks, and rather hurts the ranking. This essentially points to the effectiveness of the underlying LLM used for forward lookup (Llama-3.1-8B-Instruct for these reported results). In general, the 8B model is much worse than the 70B variant used for final generation ($\sim15\%$ lower average performance in our initial experiments). Often, the former even fails to follow our formatting instructions to generate the `Rationale:' and `Answer:' prefixes correctly. Further, we often see the answer being absent or cut off due to the model generating a long reasoning statement, leaving no room for the answer within our hard decoding token limit. However, regardless of these issues, as long as the model outputs the appropriate language relevant to answering the input question, it helps to retrieve the most relevant chunks for the final generation step by a more powerful LLM. We also experimented with different prompts for Stage II and found that some sort of reasoning or explanation provides slight gains over only using answers (Appendix \ref{sec:appendix-prompt-comparisons}).

\begin{figure}[t]
\centering
  \includegraphics[width=0.9\columnwidth]{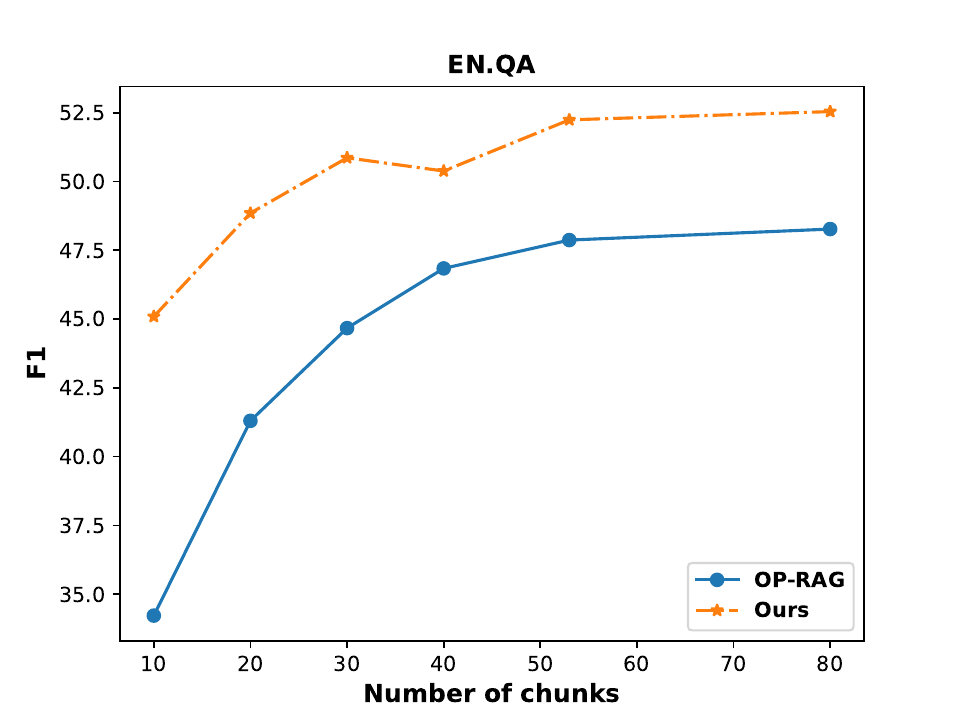}  
  \includegraphics[width=0.9\columnwidth]{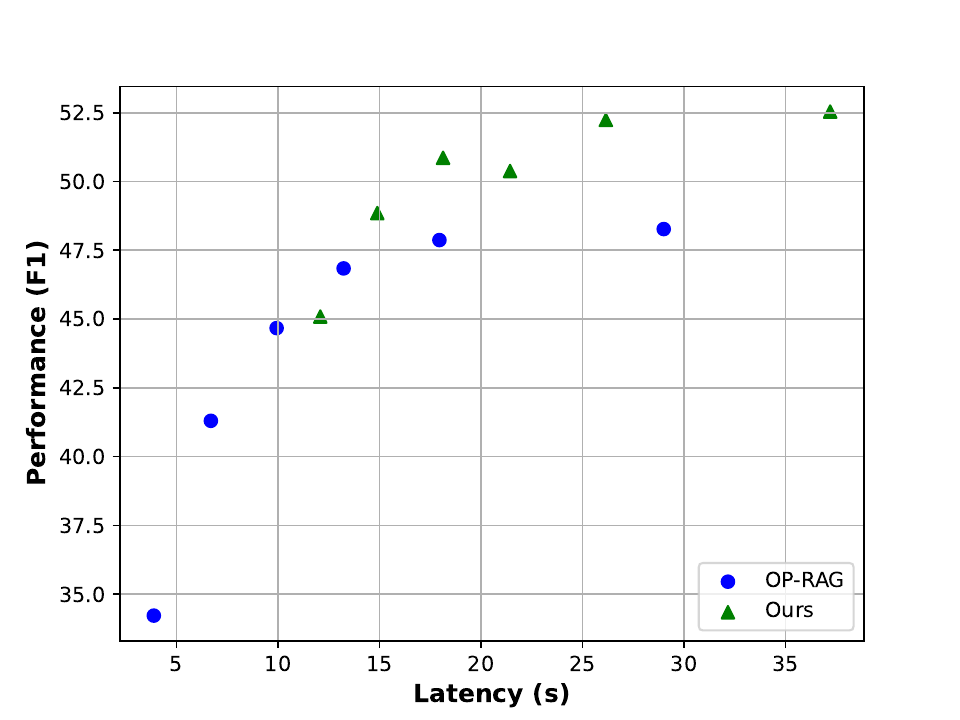}
  \caption{\textit{Top}: Results on \textbf{EN.QA} obtained by varying the number of chunks used for final response generation. Across all data points, our approach uses an Llama3.1-8B-Instruct model for forward lookup in Stage II with $80$ context chunks as input and setting $\eta_F=1$ and $\eta_B=0$. \textit{Bottom}: Performance vs. Latency plot on \textbf{EN.QA} for the same points as in the \textit{Top} Figure. Refer to Appendix \ref{sec:appendix-expt-design} for details on the hardware used.}
  \label{fig:chunks-infbench-enqa}
\end{figure}

\noindent\textbf{Forward-looking improves the ranking of relevant context chunks.} In Figure~\ref{fig:chunks-infbench-enqa} (top), we directly compare OP-RAG with our approach on \textbf{EN.QA} by varying the number of chunks used for final generation\footnote{We exclude \textbf{Self-Route} here since it relies on \textbf{LC} as a fallback which already performs poorer than RAG in this case.}. We find that our approach at $20$ chunks (6k context) outperforms OP RAG at $80$ chunks (24k context). On \textbf{EN.MC} (Appendix \ref{sec:appendix-vary-chunks-enmc}), this happens at $53$ chunks (16k context). This goes back to the discussion in Section \ref{sec:approach-overview}. With forward lookup in Stage II (albeit with a less powerful LLM), our approach essentially improves the ranking of relevant context chunks, and thus, allows one to use a smaller context for final response generation. This makes it easier for the LLM to find the correct answer, leading to improved performance.

\noindent\textbf{Performance improves even with one forward sample in Stage II of FB-RAG.} Finally, we analyze the impact of the number of samples used in Stage II of FB-RAG on the overall performance (Appendix~\ref{sec:appendix-num-samples}). We find that the performance improves greatly with only one forward sample, with maximum performance at $5$. We also note that the trend is not strictly increasing, indicating that more samples may not always add value and this parameter must be tuned empirically.

\section{Discussion}
\label{sec:discussion}

\noindent\textbf{Combining FB-RAG with In-Context Learning}: \citet{wei2024instructrag} proposed to improve the generation step in RAG with In-Context Learning (ICL) examples. This approach is complementary to \textbf{FB-RAG}, which improves the retrieval of relevant chunks. Hence, we studied how combining these two methods impacts performance. We consider two variants for the retrieval step: 1) \textbf{Vanilla} and 2) \textbf{Ours-F}). We consider three augmentations: \textit{Reason-then-Answer}, which includes reasoning in the final generation by \textbf{Llama3.1-70B-Instruct}, along with \textit{Few-Shot Demo. w/ Instruction} and \textit{INSTRUCTRAG-ICL} methods by \citet{wei2024instructrag} which combine reasoning with ICL. We implemented these methods ourselves and combined them with one of the two retrieval choices, for a total of $6$ new evaluations.

We present the results in Appendix Table \ref{tab:appendix-combine-icl}. On average, our original \textbf{Ours-F} model from Table \ref{tab:results-longbench} remains the top performer. The ICL augmentations provide a clear benefit on one dataset (\textbf{2Wiki}). Crucially, the results for the ICL-augmented \textbf{Vanilla} setting are consistently lower than for the ICL-augmented \textbf{Ours-F}. This demonstrates that while generation techniques like ICL can help, they cannot fully compensate for poor retrieval. \textbf{Ours-F} provides a better, more precise context for these generation methods to work with.

Our qualitative analysis reveals trade-offs. Adding reasoning (especially \textit{Reason-then-Answer}) often hurts instruction-following, causing the model to generate long answers instead of the concise phrases requested in the prompt. While \textit{INSTRUCTRAG-ICL} fixes some errors made by the base \textbf{Ours-F} model, its own faulty reasoning sometimes introduces new errors. One potential benefit we observed was in handling out-of-context queries: the \textit{INSTRUCTRAG-ICL} model often correctly concluded that information was missing or used its internal knowledge, whereas the base \textbf{Ours-F} model was more likely to make an incorrect guess. While this did not improve scores after the augmentations, it suggests a potential direction for improving system reliability. Effectively integrating these ICL methods remains an area for future work.

\begin{figure}[t]
\centering
  \includegraphics[width=0.9\columnwidth]{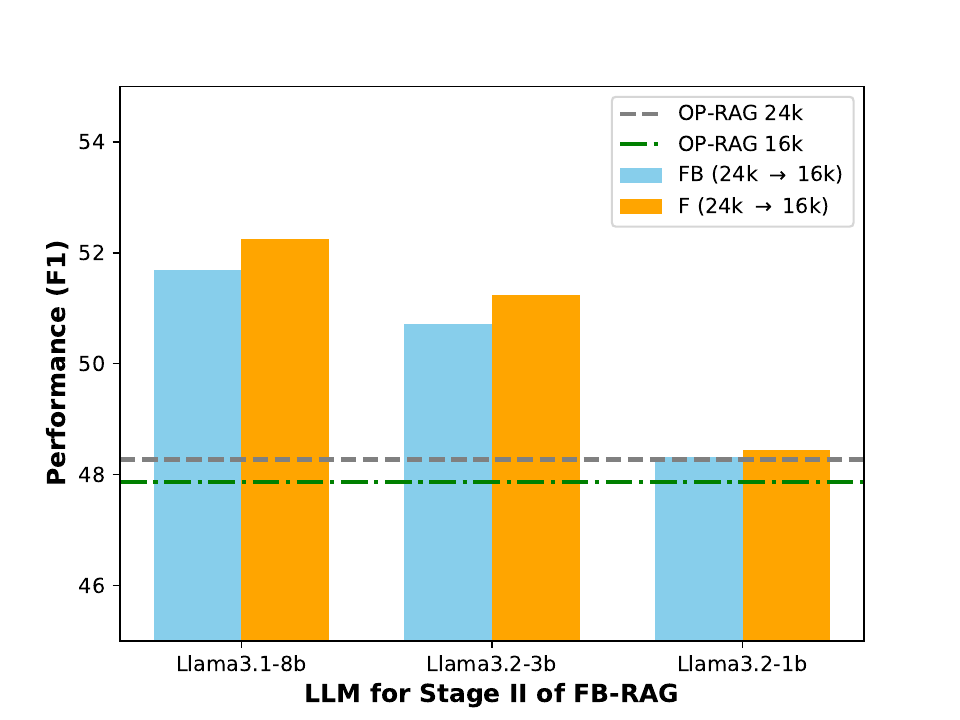}
  \caption{Varying the model used for Forward lookup in Stage II of our approach. Results are on EN.QA dataset.}
  \label{fig:light-weight}
\end{figure}

\noindent\textbf{Latency Considerations: FB-RAG improves performance with lower latency.} The latency of FB-RAG is governed by the two LLM calls in Stage II and III (Figure \ref{fig:fbrag-overview}). We approximate the overall latency by the sum of the average time taken by Llama3.1-8B-Instruct to generate output samples in Stage II (assuming parallelization) and the average time taken by Llama3.1-70B-Instruct to generate the final answer. In Figure \ref{fig:chunks-infbench-enqa} (bottom), we plot performance against latency for \textbf{EN.QA}, varying the number of chunks used in Stage III and comparing to OP-RAG. This is complementary to the performance curves in Figure \ref{fig:chunks-infbench-enqa} (top). As evident, we find that FB-RAG improves performance while reducing latency. It matches the best baseline performance ($48.27$ F1 at $29$s), with over 48\% reduction in latency, attaining $48.85$ F1 at $14.89$s. Further, FB-RAG shows 8\% performance improvement with a 10\% reduction in latency. This can be attributed to using a lightweight 8B model for forward-lookup with a large context, and the final generation with a 70B model using a much smaller context size, and is in line with previously reported inference speedups in 8B vs. 70B variants\footnote{\url{https://openllmbenchmarks.com/index.html}}.

\noindent\textbf{Varying the LLM used for forward lookup: We can go even more light-weight.} The latency analysis above used an 8B model for forward-lookup in Stage II of FB-RAG. Even though the 8B model fails to follow instructions properly occasionally and performs much worse compared to the 70B model, it still brings performance improvements. A natural question is -- `\textit{Can we push this further?}' In Figure \ref{fig:light-weight}, we compare performance by varying the LLM used for Stage II, experimenting with Llama3.2 3B and 1B instruction-tuned variants\footnote{\url{https://ai.meta.com/blog/llama-3-2-connect-2024-vision-edge-mobile-devices/}}. As evident, we find that even the 3B model shows visible improvements in performance, while the 1B performs similar to the baseline. This finding attests to the strength of FB-RAG -- although the 3B variant is nearly half as accurate as the 8B model, as long as it provides the relevant language in the generated reasons and answers, it helps to retrieve the relevant context chunks for the 70B model to generate accurate answers. From these observations, we argue that FB-RAG provides the knobs to improve performance while controlling latency with reasonable design choices -- this includes the number of chunks for Stage II and Stage III, and the size of the forward-lookup model. % We next perform qualitative analysis to look at specific examples where FB-RAG shines and where it fails, to guide future efforts in this area.

\noindent\textbf{Qualitative Analysis:} Analyzing complex queries where FB-RAG decisively outperforms the baselines, we make two observations. First (which is more straightforward), there are cases where the 8B model answers the query correctly in at least one of the Stage II samples, along with giving a reasonable rationale. This directly helps to pick the relevant chunks for Stage III following Equation \ref{eq:fb-component}. The second situation is more interesting, where the 8B model \textit{fails} to answer a multihop query in \textit{all} samples. However, it answers one hop correctly in at least one of the samples, which proves to be sufficient to retrieve the correct chunks for the 70B model to handle the multiple hops correctly. Take a query from \textbf{MuSiQue} as an example -- `\textit{Who is the spouse of the actor who played Hannibal Smith in the A team?}', the 8B model correctly guesses `\textit{George Peppard}' as the actor who played Hannibal Smith, but is unable to get to the final answer `\textit{Sherry Boucher}`. However, simply generating the relevant language and `\textit{George Peppard}' helps to retrieve the right context chunks for the 70B model to produce the correct answer -- This gives insight into how even a light-weight LLM in Stage II can systematically help to improve the performance, aligned with the overall results discussed earlier.

Looking at the fewer cases where FB-RAG performs worse, we find that first, some of the errors can be traced back to the evaluation metrics. When FB-RAG predicts `\textit{Sebastian}' instead of `\textit{Sebastian Cabot}' or `\textit{Qatari Stars League}' instead of `\textit{Qatar Stars League}', it hurts the F1 score it receives. -- \textbf{Investing in improved metrics (potentially semantic) will be valuable in the future.} Second, in some cases, the error can be attributed to the ambiguity in the input query. The answer to the question `\textit{The Live Life Loud album's band signed to which label?}' is \textit{temporally dependent}, and FB-RAG gets penalized since it answers correctly but from a different year than what is \textit{unfairly} assumed in the ground truth answer -- \textbf{Incorporating the temporal dimension to curate unambiguous queries will improve the dataset quality in the future.} Finally, we find cases where the 70B model fails to resolve multihop queries even with a precise input context, for instance, confusing the `\textit{spouse}' with the `\textit{mother}' of an artist -- \textbf{Enabling LLMs to resolve complex multihop queries is still an open, challenging problem, demanding additional dedicated efforts in this area.}

% We further performed qualitative analysis.

\section{Related Work}
\label{sec:related-work}

\noindent\textbf{Long Context (LC) LLMs}: Context lengths have rapidly increased, with Gemini $1.5$ Pro~\cite{team2024gemini} and Meta Llama $4$~\cite{llama4-2025} using even $10$ million tokens. However, LLMs can get confused by irrelevant parts of the context, leading to known cases where RAG significantly outperforms LC~\cite{xu2023retrieval,yu2024defense}. In terms of latency, LC is expensive due to the quadratically increasing compute costs with input size. We follow the RAG paradigm by first retrieving the most relevant context chunks and then feeding them to an LLM with the input query for answer generation.

\noindent\textbf{Retrieval Augmented Generation (RAG)}: RAG has emerged as a popular paradigm competing with LC, improving performance across diverse tasks with significantly lower compute costs~\cite{fan2024survey}. Traditional RAG is \textit{backward-looking} -- the context chunks are scored based on the input query using a combination of retrievers and rerankers, which further refine the selected context~\cite{gao2023retrieval}. Instead, FB-RAG uses \textit{forward-looking} with samples generated from an LLM to select the relevant context chunks for the final answer generation. Unlike a typical reranker, Stage II of FB-RAG selects the chunks from the full context $C$ instead of $C^{I}$ (the output of Stage 1).

% Other efforts have explored look-ahead ideas to leverage model uncertainty in related but different contexts. For instance, both   LLM's confidence to retrieve from the web. FLARE  Cite FLARE: https://arxiv.org/pdf/2305.06983 

Numerous efforts augment RAG with trained filters~\cite{yoran2023making}, trained compressors~\cite{xu2024recomp}, and web search engines~\cite{yan2024corrective} to improve retrieval quality and generation. Self-RAG~\cite{asai2024self} trains an LLM using special reflection tokens to retrieve on demand. ~\citet{li2023web} and ~\citet{jiang-etal-2023-active} perform retrieval from the web based on the LLM's look-ahead confidence scores. Speculative RAG~\cite{wang2024speculative} uses a smaller trained LLM to generate answer candidates, which are then verified by another LLM. LongRAG~\cite{zhao2024longrag} uses plug-n-play components to extract global information and factual details from context chunks which enhances the understanding from long contexts. Our setting differs in several ways: 1) We push the performance of instruction-tuned LLMs without any further training, 2) We assume no access to external web sources, and 3) We only use forward lookup from the light-weight LLM in a \textit{soft} manner for selecting relevant context chunks from the entire context, with the final generation still being performed by a more powerful LLM. Two recent papers closest to our formulation are Self-Route~\cite{li-etal-2024-retrieval} and Order Preserving (OP) RAG~\cite{yu2024defense}, which we implemented ourselves and used as baselines in this work.

\section{Conclusion}
We proposed and evaluated FB-RAG -- a new framework for RAG with LLMs. Instead of solely relying on the input query to retrieve the relevant chunks, we employed a look-ahead mechanism tightly integrated with the task at hand. This retrieves the most relevant chunks while reducing the irrelevant information in the context, resulting in superior performance. We found that FB-RAG has the potential to improve performance while simultaneously reducing latency. We performed a qualitative analysis and discussed insights to guide future work.

Our findings also provide clear guidance on when to use FB-RAG. For applications with similar setup as ours, \textbf{Ours-F} (forward-lookup only) should be preferred over \textbf{Ours-FB} (forward and backward) given its superior performance in our experiments. The choice between \textbf{Ours-F} and \textbf{OP-RAG} depends on the specific performance versus latency requirements:
\begin{itemize}
    \item \textbf{Performance-first (e.g., offline setup):} As shown in Table \ref{tab:results-longbench} and \ref{tab:results-infbench} along with Figure \ref{fig:chunks-infbench-enqa} (top), \textbf{Ours-F} consistently achieves the best performance.
    \item \textbf{Fixed performance target:} As shown in Figure \ref{fig:chunks-infbench-enqa} (bottom), \textbf{OP-RAG} may be suitable for lower performance targets, but for higher targets, \textbf{Ours-F} achieves them with lower latency.
    \item \textbf{Fixed latency budget:} For low latency budgets, \textbf{OP-RAG} is preferable, but as the budget increases, \textbf{Ours-F} delivers superior performance.
\end{itemize}
This highlights a key advantage of \textbf{FB-RAG} for long-text applications. When the input context is large (e.g., 16k or 24k), standard RAG with a large generator (e.g., 70B model) incurs high latency. In such cases, \textbf{FB-RAG}'s strategy of using a light-weight model for forward-lookup allows it to achieve higher performance while simultaneously reducing overall latency.

\section*{Limitations}
The effectiveness of \textbf{FB-RAG} depends on the quality of the off-the-shelf retriever that is being used. In our experiments, we found BM25 to be effective. However, if the quality of the available retriever is poor for the domain in consideration, one is forced to use a larger context size for subsequent stages, which would impact the overall latency gains from the system.

In our qualitative analysis, we observed that smaller, less-capable LLMs can be used for forward-lookup in Stage II and one can eventually get accurate responses even if this small LLM is inaccurate or fails to always follow our instructions properly. However, the minimum level of the model capability (parameters, instruction-following abilities) required for forward-looking signals to be helpful remains an open question and will be important for future investigation. Ultimately, the design choices that best manage the performance-latency tradeoff will depend on the specific application and platform constraints.

% First, note that our focus in this work has been on closed-form generation, meaning that we assumed access to an input context that is sufficient for the model to generate the answer. While out of scope from our current investigation, it can be useful to extend our method to formulations that allow access to external knowledge sources and web search engines. This can potentially handle a broader set of input queries, alleviating the need to curate an input context ahead of time.

% Furthermore, while our objective was to improve the performance of instruction-tuned LLMs without any additional fine-tuning, this can be explored to further improve performance in the presence of relevant domain-specific training data.
% limitations go here

\section*{Ethical Considerations}

Our work was approved by the established internal review procedure. We carefully verified the licensing information associated with all the datasets and instruction-tuned LLMs used in this work, ensuring that their use was within their intended scope. All the datasets were properly anonymized before being used. We provide dataset statistics in Appendix \ref{sec:appendix-datasets} and refer the readers to the original dataset papers for details regarding pre-processing steps as well as the demographics of human annotators.

All datasets considered in this work were in English. Hence, it is unclear whether our findings directly translate to other languages and cultures. However, our approach is free of any such assumptions, and we encourage future work to extend it to these other scenarios.

We further note that LLMs have been known to exhibit different kinds of gender or cultural biases that can lead to discriminatory language in the generated outputs. Hence, we call for rigorous testing before any LLM-based systems are deployed. We also recommend regular monitoring after deployment to ensure that the models' behaviors remain within their planned scope.

% All the datasets used in this work 

% internal approval was done.
% llms and datasts - proper licensing and carefully verified that the use is within the intended scope.

% datasets properly anonymized

% llms biases - proper testing before deployment and regular monitoring after.

% all in english so unclear -> but the approach is generic and potentially be experimented.

% datasets -- proper license was followed. llms are prone to biases uses off-the-shelf..take language from the papers of these models like llama and cite them. so proper monitoring and caution before deployment.
% we provide details for reproducability which will help.

% Bibliography entries for the entire Anthology, followed by custom entries
%\bibliography{anthology,custom}
% Custom bibliography entries only
\bibliography{custom, anthology}

\begin{thebibliography}{37}
\providecommand{\natexlab}[1]{#1}

\bibitem[{Asai et~al.(2024)Asai, Wu, Wang, Sil, and Hajishirzi}]{asai2024self}
Akari Asai, Zeqiu Wu, Yizhong Wang, Avi Sil, and Hannaneh Hajishirzi. 2024.
\newblock Self-rag: Learning to retrieve, generate, and critique through self-reflection.
\newblock In \emph{International Conference on Learning Representations}.

\bibitem[{Bai et~al.(2024)Bai, Lv, Zhang, Lyu, Tang, Huang, Du, Liu, Zeng, Hou, Dong, Tang, and Li}]{bai-etal-2024-longbench}
Yushi Bai, Xin Lv, Jiajie Zhang, Hongchang Lyu, Jiankai Tang, Zhidian Huang, Zhengxiao Du, Xiao Liu, Aohan Zeng, Lei Hou, Yuxiao Dong, Jie Tang, and Juanzi Li. 2024.
\newblock \href {https://doi.org/10.18653/v1/2024.acl-long.172} {{L}ong{B}ench: A bilingual, multitask benchmark for long context understanding}.
\newblock In \emph{Proceedings of the 62nd Annual Meeting of the Association for Computational Linguistics (Volume 1: Long Papers)}, pages 3119--3137, Bangkok, Thailand. Association for Computational Linguistics.

\bibitem[{Borgeaud et~al.(2022)Borgeaud, Mensch, Hoffmann, Cai, Rutherford, Millican, Van Den~Driessche, Lespiau, Damoc, Clark et~al.}]{borgeaud2022improving}
Sebastian Borgeaud, Arthur Mensch, Jordan Hoffmann, Trevor Cai, Eliza Rutherford, Katie Millican, George~Bm Van Den~Driessche, Jean-Baptiste Lespiau, Bogdan Damoc, Aidan Clark, and 1 others. 2022.
\newblock Improving language models by retrieving from trillions of tokens.
\newblock In \emph{International conference on machine learning}, pages 2206--2240. PMLR.

\bibitem[{Chen et~al.(2024)Chen, Xiao, Zhang, Luo, Lian, and Liu}]{chen-etal-2024-m3}
Jianlyu Chen, Shitao Xiao, Peitian Zhang, Kun Luo, Defu Lian, and Zheng Liu. 2024.
\newblock \href {https://doi.org/10.18653/v1/2024.findings-acl.137} {{M}3-embedding: Multi-linguality, multi-functionality, multi-granularity text embeddings through self-knowledge distillation}.
\newblock In \emph{Findings of the Association for Computational Linguistics: ACL 2024}, pages 2318--2335, Bangkok, Thailand. Association for Computational Linguistics.

\bibitem[{Dasigi et~al.(2021)Dasigi, Lo, Beltagy, Cohan, Smith, and Gardner}]{dasigi-etal-2021-dataset}
Pradeep Dasigi, Kyle Lo, Iz~Beltagy, Arman Cohan, Noah~A. Smith, and Matt Gardner. 2021.
\newblock \href {https://doi.org/10.18653/v1/2021.naacl-main.365} {A dataset of information-seeking questions and answers anchored in research papers}.
\newblock In \emph{Proceedings of the 2021 Conference of the North American Chapter of the Association for Computational Linguistics: Human Language Technologies}, pages 4599--4610, Online. Association for Computational Linguistics.

\bibitem[{Fan et~al.(2024)Fan, Ding, Ning, Wang, Li, Yin, Chua, and Li}]{fan2024survey}
Wenqi Fan, Yujuan Ding, Liangbo Ning, Shijie Wang, Hengyun Li, Dawei Yin, Tat-Seng Chua, and Qing Li. 2024.
\newblock A survey on rag meeting llms: Towards retrieval-augmented large language models.
\newblock In \emph{Proceedings of the 30th ACM SIGKDD Conference on Knowledge Discovery and Data Mining}, pages 6491--6501.

\bibitem[{Gao et~al.(2023)Gao, Xiong, Gao, Jia, Pan, Bi, Dai, Sun, Wang, and Wang}]{gao2023retrieval}
Yunfan Gao, Yun Xiong, Xinyu Gao, Kangxiang Jia, Jinliu Pan, Yuxi Bi, Yi~Dai, Jiawei Sun, Haofen Wang, and Haofen Wang. 2023.
\newblock Retrieval-augmented generation for large language models: A survey.
\newblock \emph{arXiv preprint arXiv:2312.10997}, 2.

\bibitem[{Guu et~al.(2020)Guu, Lee, Tung, Pasupat, and Chang}]{guu2020retrieval}
Kelvin Guu, Kenton Lee, Zora Tung, Panupong Pasupat, and Mingwei Chang. 2020.
\newblock Retrieval augmented language model pre-training.
\newblock In \emph{International conference on machine learning}, pages 3929--3938. PMLR.

\bibitem[{He et~al.(2021)He, Huang, Cui, Li, and Liu}]{he2021fast}
Qiuxiang He, Guoping Huang, Qu~Cui, Li~Li, and Lemao Liu. 2021.
\newblock Fast and accurate neural machine translation with translation memory.
\newblock In \emph{Proceedings of the 59th Annual Meeting of the Association for Computational Linguistics and the 11th International Joint Conference on Natural Language Processing (Volume 1: Long Papers)}, pages 3170--3180.

\bibitem[{Ho et~al.(2020)Ho, Nguyen, Sugawara, and Aizawa}]{ho2020constructing}
Xanh Ho, Anh-Khoa~Duong Nguyen, Saku Sugawara, and Akiko Aizawa. 2020.
\newblock Constructing a multi-hop qa dataset for comprehensive evaluation of reasoning steps.
\newblock In \emph{Proceedings of the 28th International Conference on Computational Linguistics}, pages 6609--6625.

\bibitem[{Jiang et~al.(2023)Jiang, Xu, Gao, Sun, Liu, Dwivedi-Yu, Yang, Callan, and Neubig}]{jiang-etal-2023-active}
Zhengbao Jiang, Frank Xu, Luyu Gao, Zhiqing Sun, Qian Liu, Jane Dwivedi-Yu, Yiming Yang, Jamie Callan, and Graham Neubig. 2023.
\newblock \href {https://doi.org/10.18653/v1/2023.emnlp-main.495} {Active retrieval augmented generation}.
\newblock In \emph{Proceedings of the 2023 Conference on Empirical Methods in Natural Language Processing}, pages 7969--7992, Singapore. Association for Computational Linguistics.

\bibitem[{Khandelwal et~al.(2019)Khandelwal, Levy, Jurafsky, Zettlemoyer, and Lewis}]{khandelwalgeneralization}
Urvashi Khandelwal, Omer Levy, Dan Jurafsky, Luke Zettlemoyer, and Mike Lewis. 2019.
\newblock Generalization through memorization: Nearest neighbor language models.
\newblock In \emph{International Conference on Learning Representations}.

\bibitem[{Ko{\v{c}}isk{\'y} et~al.(2018)Ko{\v{c}}isk{\'y}, Schwarz, Blunsom, Dyer, Hermann, Melis, and Grefenstette}]{kocisky-etal-2018-narrativeqa}
Tom{\'a}{\v{s}} Ko{\v{c}}isk{\'y}, Jonathan Schwarz, Phil Blunsom, Chris Dyer, Karl~Moritz Hermann, G{\'a}bor Melis, and Edward Grefenstette. 2018.
\newblock \href {https://doi.org/10.1162/tacl_a_00023} {The {N}arrative{QA} reading comprehension challenge}.
\newblock \emph{Transactions of the Association for Computational Linguistics}, 6:317--328.

\bibitem[{Li et~al.(2023)Li, Tang, Zhao, Wang, Nie, and Wen}]{li2023web}
Junyi Li, Tianyi Tang, Wayne~Xin Zhao, Jingyuan Wang, Jian-Yun Nie, and Ji-Rong Wen. 2023.
\newblock The web can be your oyster for improving large language models.
\newblock \emph{arXiv preprint arXiv:2305.10998}.

\bibitem[{Li et~al.(2024)Li, Li, Zhang, Mei, and Bendersky}]{li-etal-2024-retrieval}
Zhuowan Li, Cheng Li, Mingyang Zhang, Qiaozhu Mei, and Michael Bendersky. 2024.
\newblock \href {https://doi.org/10.18653/v1/2024.emnlp-industry.66} {Retrieval augmented generation or long-context {LLM}s? a comprehensive study and hybrid approach}.
\newblock In \emph{Proceedings of the 2024 Conference on Empirical Methods in Natural Language Processing: Industry Track}, pages 881--893, Miami, Florida, US. Association for Computational Linguistics.

\bibitem[{Liu et~al.(2023)Liu, Nie, Wang, Lu, Qiao, Liu, Tang, Xiao, and Anandkumar}]{liu2023multi}
Shengchao Liu, Weili Nie, Chengpeng Wang, Jiarui Lu, Zhuoran Qiao, Ling Liu, Jian Tang, Chaowei Xiao, and Animashree Anandkumar. 2023.
\newblock Multi-modal molecule structure--text model for text-based retrieval and editing.
\newblock \emph{Nature Machine Intelligence}, 5(12):1447--1457.

\bibitem[{Meta(2024)}]{llama31-2024}
Meta. 2024.
\newblock \href {https://www.llama.com/llama3_1/} {[link]}.

\bibitem[{Meta(2025)}]{llama4-2025}
Meta. 2025.
\newblock \href {https://ai.meta.com/blog/llama-4-multimodal-intelligence/} {[link]}.

\bibitem[{Sun et~al.(2022)Sun, Wang, Tay, Yang, and Zhou}]{sun2022recitation}
Zhiqing Sun, Xuezhi Wang, Yi~Tay, Yiming Yang, and Denny Zhou. 2022.
\newblock Recitation-augmented language models.
\newblock \emph{arXiv preprint arXiv:2210.01296}.

\bibitem[{Team et~al.(2024)Team, Georgiev, Lei, Burnell, Bai, Gulati, Tanzer, Vincent, Pan, Wang et~al.}]{team2024gemini}
Gemini Team, Petko Georgiev, Ving~Ian Lei, Ryan Burnell, Libin Bai, Anmol Gulati, Garrett Tanzer, Damien Vincent, Zhufeng Pan, Shibo Wang, and 1 others. 2024.
\newblock Gemini 1.5: Unlocking multimodal understanding across millions of tokens of context.
\newblock \emph{arXiv preprint arXiv:2403.05530}.

\bibitem[{Trivedi et~al.(2022)Trivedi, Balasubramanian, Khot, and Sabharwal}]{trivedi2022musique}
Harsh Trivedi, Niranjan Balasubramanian, Tushar Khot, and Ashish Sabharwal. 2022.
\newblock Musique: Multihop questions via single-hop question composition.
\newblock \emph{Transactions of the Association for Computational Linguistics}, 10:539--554.

\bibitem[{Trotman et~al.(2014)Trotman, Puurula, and Burgess}]{trotman2014improvements}
Andrew Trotman, Antti Puurula, and Blake Burgess. 2014.
\newblock Improvements to bm25 and language models examined.
\newblock In \emph{Proceedings of the 19th Australasian Document Computing Symposium}, pages 58--65.

\bibitem[{Wang et~al.(2024)Wang, Wang, Le, Zheng, Mishra, Perot, Zhang, Mattapalli, Taly, Shang et~al.}]{wang2024speculative}
Zilong Wang, Zifeng Wang, Long Le, Huaixiu~Steven Zheng, Swaroop Mishra, Vincent Perot, Yuwei Zhang, Anush Mattapalli, Ankur Taly, Jingbo Shang, and 1 others. 2024.
\newblock Speculative rag: Enhancing retrieval augmented generation through drafting.
\newblock \emph{arXiv preprint arXiv:2407.08223}.

\bibitem[{Wei et~al.(2024)Wei, Chen, and Meng}]{wei2024instructrag}
Zhepei Wei, Wei-Lin Chen, and Yu~Meng. 2024.
\newblock Instructrag: Instructing retrieval-augmented generation via self-synthesized rationales.
\newblock \emph{arXiv preprint arXiv:2406.13629}.

\bibitem[{Wu et~al.(2024)Wu, Chang, Yu, He, Wang, Hou, and McAuley}]{wu2024coral}
Junda Wu, Cheng-Chun Chang, Tong Yu, Zhankui He, Jianing Wang, Yupeng Hou, and Julian McAuley. 2024.
\newblock Coral: Collaborative retrieval-augmented large language models improve long-tail recommendation.
\newblock In \emph{Proceedings of the 30th ACM SIGKDD Conference on Knowledge Discovery and Data Mining}, pages 3391--3401.

\bibitem[{Xiao et~al.(2024)Xiao, Liu, Zhang, Muennighoff, Lian, and Nie}]{xiao2024c}
Shitao Xiao, Zheng Liu, Peitian Zhang, Niklas Muennighoff, Defu Lian, and Jian-Yun Nie. 2024.
\newblock C-pack: Packed resources for general chinese embeddings.
\newblock In \emph{Proceedings of the 47th international ACM SIGIR conference on research and development in information retrieval}, pages 641--649.

\bibitem[{Xu et~al.(2024)Xu, Shi, and Choi}]{xu2024recomp}
Fangyuan Xu, Weijia Shi, and Eunsol Choi. 2024.
\newblock Recomp: Improving retrieval-augmented lms with compression and selective augmentation.
\newblock In \emph{12th International Conference on Learning Representations, ICLR 2024}.

\bibitem[{Xu et~al.(2023)Xu, Ping, Wu, McAfee, Zhu, Liu, Subramanian, Bakhturina, Shoeybi, and Catanzaro}]{xu2023retrieval}
Peng Xu, Wei Ping, Xianchao Wu, Lawrence McAfee, Chen Zhu, Zihan Liu, Sandeep Subramanian, Evelina Bakhturina, Mohammad Shoeybi, and Bryan Catanzaro. 2023.
\newblock Retrieval meets long context large language models.
\newblock In \emph{The Twelfth International Conference on Learning Representations}.

\bibitem[{Yan et~al.(2024)Yan, Gu, Zhu, and Ling}]{yan2024corrective}
Shi-Qi Yan, Jia-Chen Gu, Yun Zhu, and Zhen-Hua Ling. 2024.
\newblock Corrective retrieval augmented generation.
\newblock \emph{arXiv preprint arXiv:2401.15884}.

\bibitem[{Yang et~al.(2023)Yang, Li, Zhang, Wang, Cheng, Li, and Xiao}]{yang2023prca}
Haoyan Yang, Zhitao Li, Yong Zhang, Jianzong Wang, Ning Cheng, Ming Li, and Jing Xiao. 2023.
\newblock Prca: Fitting black-box large language models for retrieval question answering via pluggable reward-driven contextual adapter.
\newblock In \emph{Proceedings of the 2023 Conference on Empirical Methods in Natural Language Processing}, pages 5364--5375.

\bibitem[{Yang et~al.(2018)Yang, Qi, Zhang, Bengio, Cohen, Salakhutdinov, and Manning}]{yang-etal-2018-hotpotqa}
Zhilin Yang, Peng Qi, Saizheng Zhang, Yoshua Bengio, William Cohen, Ruslan Salakhutdinov, and Christopher~D. Manning. 2018.
\newblock \href {https://doi.org/10.18653/v1/D18-1259} {{H}otpot{QA}: A dataset for diverse, explainable multi-hop question answering}.
\newblock In \emph{Proceedings of the 2018 Conference on Empirical Methods in Natural Language Processing}, pages 2369--2380, Brussels, Belgium. Association for Computational Linguistics.

\bibitem[{Yoran et~al.(2023)Yoran, Wolfson, Ram, and Berant}]{yoran2023making}
Ori Yoran, Tomer Wolfson, Ori Ram, and Jonathan Berant. 2023.
\newblock Making retrieval-augmented language models robust to irrelevant context.
\newblock \emph{arXiv preprint arXiv:2310.01558}.

\bibitem[{Yu et~al.(2024)Yu, Xu, and Akkiraju}]{yu2024defense}
Tan Yu, Anbang Xu, and Rama Akkiraju. 2024.
\newblock In defense of rag in the era of long-context language models.
\newblock \emph{arXiv preprint arXiv:2409.01666}.

\bibitem[{Zhang et~al.(2024)Zhang, Chen, Hu, Xu, Chen, Hao, Han, Thai, Wang, Liu, and Sun}]{zhang-etal-2024-bench}
Xinrong Zhang, Yingfa Chen, Shengding Hu, Zihang Xu, Junhao Chen, Moo Hao, Xu~Han, Zhen Thai, Shuo Wang, Zhiyuan Liu, and Maosong Sun. 2024.
\newblock \href {https://doi.org/10.18653/v1/2024.acl-long.814} {$\infty${B}ench: Extending long context evaluation beyond 100{K} tokens}.
\newblock In \emph{Proceedings of the 62nd Annual Meeting of the Association for Computational Linguistics (Volume 1: Long Papers)}, pages 15262--15277, Bangkok, Thailand. Association for Computational Linguistics.

\bibitem[{Zhao et~al.(2024{\natexlab{a}})Zhao, Wang, Cen, Zha, Tan, Dong, and Tang}]{zhao2024longrag}
Qingfei Zhao, Ruobing Wang, Yukuo Cen, Daren Zha, Shicheng Tan, Yuxiao Dong, and Jie Tang. 2024{\natexlab{a}}.
\newblock Longrag: A dual-perspective retrieval-augmented generation paradigm for long-context question answering.
\newblock In \emph{Proceedings of the 2024 Conference on Empirical Methods in Natural Language Processing}, pages 22600--22632.

\bibitem[{Zhao et~al.(2024{\natexlab{b}})Zhao, Yang, Wang, He, Qiu, and Qiu}]{zhao2024retrieval}
Siyun Zhao, Yuqing Yang, Zilong Wang, Zhiyuan He, Luna~K Qiu, and Lili Qiu. 2024{\natexlab{b}}.
\newblock Retrieval augmented generation (rag) and beyond: A comprehensive survey on how to make your llms use external data more wisely.
\newblock \emph{arXiv preprint arXiv:2409.14924}.

\bibitem[{Zhong et~al.(2021)Zhong, Yin, Yu, Zaidi, Mutuma, Jha, Awadallah, Celikyilmaz, Liu, Qiu, and Radev}]{zhong-etal-2021-qmsum}
Ming Zhong, Da~Yin, Tao Yu, Ahmad Zaidi, Mutethia Mutuma, Rahul Jha, Ahmed~Hassan Awadallah, Asli Celikyilmaz, Yang Liu, Xipeng Qiu, and Dragomir Radev. 2021.
\newblock \href {https://doi.org/10.18653/v1/2021.naacl-main.472} {{QMS}um: A new benchmark for query-based multi-domain meeting summarization}.
\newblock In \emph{Proceedings of the 2021 Conference of the North American Chapter of the Association for Computational Linguistics: Human Language Technologies}, pages 5905--5921, Online. Association for Computational Linguistics.

\end{thebibliography}

\appendix

\section{Methodology}
\label{sec:appendix-methodology}

In this Section, we provide a deeper insight into how FB-RAG works to improve the overall RAG performance. This interpretation is complementary to the discussion in Section \ref{sec:approach-overview}. We lay out a probabilistic formulation of the RAG process below (extending the notation used in the main paper):

\begin{equation}
    P(A^* | Q, C) = \sum_{\forall r \subseteq C} P(r | Q) \cdot P(A^* | Q, r),
\end{equation}
where $r$ denotes all possible contexts that can be selected in the retriever stage of RAG.

The first component, $P(r | Q)$, captures the retriever's role - a conditional probability distribution over all possible contexts that can be selected from the full context $C$ conditioned on the query $Q$. A higher probability of a specific $r$ corresponds to a higher score from the retriever and a higher likelihood of it being picked up for generation.

The second component, $P(A^* | Q, r)$,  captures the job of the generator - the probability of generating the answer $A^*$ from the query $Q$ and the selected context $r$. Note that $P(A^* | Q, r)$ will be high for a better quality $r$ which contains the relevant context chunks and minimizes irrelevant information, and will be low for a poor quality $r$ which misses out key relevant chunks or contains a high amount of irrelevant content.

Under this formulation, when supplied with a reasonable forward-looking LLM, the procedure laid out in Section \ref{sec:fb-retriever} simply works to \textit{shift the probability mass} in $P(r | Q)$ to better quality contexts. Combined with the better performance from the generator $P(A^* | Q, r)$ for these better quality contexts, this holds the potential to improve the overall probability $P(A^* | Q, C)$ of generating the accurate answer.

\section{Datasets}
\label{sec:appendix-datasets}
Our experiments are based on $9$ datasets from two popular benchmarks consisting long context lengths - LongBench~\cite{bai-etal-2024-longbench} and $\infty$Bench~\cite{zhang-etal-2024-bench}. QA tasks (\textit{NarrativeQA}, \textit{Qasper}, \textit{MultifieldQA}, \textit{HotpotQA}, \textit{2WikiMultihopQA}, \textit{MuSiQue}, and \textit{EN.QA}) take a query and a context as input, with the goal of generating a concise answer. The summarization task (\textit{QMSum}) requires generating a free-form summary based on the given query and context. For the MCQ task (\textit{EN.MC}), the input additionally includes a set of choices, and the task is to choose the correct choice that answers the input query based on the provided context. We present key statistics for these datasets in Table \ref{tab:data-stats}.

\begin{table}
  \centering
  \scalebox{0.8}{
  \begin{tabular}{lcc}
    \hline
    \textbf{Dataset} & \textbf{No. of Queries} &  \textbf{Avg Length} \\
    \hline
    \multicolumn{3}{c}{LongBench~\cite{bai-etal-2024-longbench}} \\
    NarrativeQA & 200 & 18,395\\
    Qasper & 200 & 3,599\\
    MultiFieldQA & 150 & 4,539\\
    HotpotQA & 200 & 9,133\\
    2WikiMultihopQA & 200 & 4,873\\
    MuSiQue & 200 & 11,196\\
    QMSum & 200 & 10,533\\
    \multicolumn{3}{c}{$\infty$Bench~\cite{zhang-etal-2024-bench}} \\
    EN.QA & 351 & 150,374\\
    EN.MC & 229 & 142,622\\ \hline
  \end{tabular}}
  \caption{Statistics for all the datasets considered in our experiments in this paper.}
  \label{tab:data-stats}
\end{table}

\section{Experiment Design}
\label{sec:appendix-expt-design}

We provide additional experimental design details in this section to promote reproducibility. %We further plan to release our code on acceptance.

\subsection{Prompts}

We release all the prompts used in our experiments. Tables \ref{tab:prompts-lb-1} and \ref{tab:prompts-lb-2} list the prompts for LongBench datasets, while Table \ref{tab:prompts-ib} presents the prompts for the two datasets from $\infty$Bench. Note that for \textbf{QMSum}, we use the same prompt for FB-RAG Stage II as the one used for Vanilla RAG. This is because the output summary is already descriptive, unlike other datasets where answers tend to be very concise (a few words or a phrase).

\subsection{Hardware Used}

All the experiments presented in this paper were performed on $8$ NVIDIA A100 GPUs. We used the default inference configuration provided by Huggingface, which uses `\textit{device\_map=auto}'. We did not use any additional optimizations.

\subsection{Decoding Token Limits}

We set maximum limits for the number of tokens that can be generated per LLM call. For LongBench datasets, we use the limits from the code released with the benchmark\footnote{\url{https://github.com/THUDM/LongBench/tree/main}}. For EN.QA and EN.MC datasets from $\infty$Bench benchmark, we set the limit to $64$, based on the ground truth distributions. When generating both reasoning and answer in Stage II of our approach, we add $64$ to the original token limit for all datasets.

% \noindent\textbf{Maximum context token limits}: Our chunking methodology is based on word-level. We additionally set maximum token limits for the context. This was set to $15,500$ for RAG experiments with $LongBench$ datasets and a limit of $48,000$ for $\Infty$Bench datasets.

\begin{table*}
  \centering
  \scalebox{0.8}{
  \begin{tabular}{l|p{4cm}|p{5cm}|p{4cm}}
    \hline
    \textbf{Dataset} & \textbf{LC, Vanilla / OP RAG} & \textbf{Self-Route: Stage I} & \textbf{FB-RAG: Stage II} \\ \hline
    NarrativeQA & \small You are given a story, which can be either a novel or a movie script, and a question. Answer the question as concisely as you can, using a single phrase if possible. Do not provide any explanation. Story: \{context\} Now, answer the question based on the story as concisely as you can, using a single phrase if possible. Do not provide any explanation. Question: \{input\} Answer: & \small You are given a story, which can be either a novel or a movie script, and a question. Answer the question as concisely as you can, using a single phrase if possible. Do not provide any explanation. If the question cannot be answered based on the information in the article, write “unanswerable”. Story: \{context\} Now, answer the question based on the story as concisely as you can, using a single phrase if possible. Do not provide any explanation. If the question cannot be answered based on the information in the article, write “unanswerable”. Question: \{input\} Answer: & \small You are given a story, which can be either a novel or a movie script, and a question. Answer the question as concisely as you can, using a single phrase if possible. Story: \{context\} Now, first provide your reasoning briefly in 2-3 sentences starting with 'Rationale:'. Then, answer the question starting with 'Answer:' as concisely as you can, using a single phrase if possible. Question: \{input\} Rationale:\\ \hline
    Qasper & \small You are given a scientific article and a question. Answer the question as concisely as you can, using a single phrase or sentence if possible. If the question cannot be answered based on the information in the article, write \"unanswerable\". If the question is a yes/no question, answer \"yes\", \"no\", or \"unanswerable\". Do not provide any explanation. Article: \{context\} Answer the question based on the above article as concisely as you can, using a single phrase or sentence if possible. If the question cannot be answered based on the information in the article, write \"unanswerable\". If the question is a yes/no question, answer \"yes\", \"no\", or \"unanswerable\". Do not provide any explanation. Question: \{input\} Answer: & \small You are given a scientific article and a question. Answer the question as concisely as you can, using a single phrase or sentence if possible. If the question cannot be answered based on the information in the article, write \"unanswerable\". If the question is a yes/no question, answer \"yes\", \"no\", or \"unanswerable\". Do not provide any explanation. Article: \{context\} Answer the question based on the above article as concisely as you can, using a single phrase or sentence if possible. If the question cannot be answered based on the information in the article, write \"unanswerable\". If the question is a yes/no question, answer \"yes\", \"no\", or \"unanswerable\". Do not provide any explanation. Question: \{input\} Answer: & \small You are given a scientific article and a question. Answer the question as concisely as you can, using a single phrase or sentence if possible. If the question cannot be answered based on the information in the article, write \"unanswerable\". If the question is a yes/no question, answer \"yes\", \"no\", or \"unanswerable\". Article: \{context\} Now, first provide your reasoning briefly in 2-3 sentences starting with 'Rationale:'. Then, answer the question starting with 'Answer:' based on the above article as concisely as you can, using a single phrase or sentence if possible. If the question cannot be answered based on the information in the article, write \"unanswerable\". If the question is a yes/no question, answer \"yes\", \"no\", or \"unanswerable\". Question: \{input\} Rationale:\\ \hline
    MultiFieldQA & \small Read the following text and answer briefly. \{context\} Now, answer the following question based on the above text, only give me the answer and do not output any other words. Question: \{input\}  Answer: & \small Read the following text and answer briefly. \{context\} Now, answer the following question based on the above text, only give me the answer and do not output any other words. If the question cannot be answered based on the information in the article, write “unanswerable”. Question: \{input\} Answer: & \small Read the following text and answer briefly. \{context\} Now, first provide your reasoning briefly in 2-3 sentences starting with 'Rationale:'. Then, answer the question starting with 'Answer:' based on the above text. Question: \{input\} Rationale:\\ \hline
    HotpotQA & \small Answer the question based on the given passages. Only give me the answer and do not output any other words. The following are given passages. \{context\} Answer the question based on the given passages. Only give me the answer and do not output any other words. Question: \{input\} Answer: & \small Answer the question based on the given passages. Only give me the answer and do not output any other words. If the question cannot be answered based on the information in the article, write “unanswerable”. The following are given passages. \{context\} Answer the question based on the given passages. Only give me the answer and do not output any other words. If the question cannot be answered based on the information in the article, write “unanswerable”. Question: \{input\} Answer: & \small Answer the question based on the given passages. \{context\} Now, first provide your reasoning briefly in 2-3 sentences starting with 'Rationale:'. Then, answer the question starting with 'Answer:' based on the given passages. Question: \{input\} Rationale:\\ \hline
  \end{tabular}}
  \caption{(Part 1 / 2) Prompts used in our experiments for LongBench datasets.}
  \label{tab:prompts-lb-1}
\end{table*}

\begin{table*}
  \centering
  \scalebox{0.8}{
  \begin{tabular}{l|p{4cm}|p{5cm}|p{4cm}}
    \hline
    \textbf{Dataset} & \textbf{LC, Vanilla / OP RAG} & \textbf{Self-Route: Stage I} & \textbf{FB-RAG: Stage II} \\ \hline
    2WikiMultihopQA & \small Answer the question based on the given passages. Only give me the answer and do not output any other words. The following are given passages. \{context\} Answer the question based on the given passages. Only give me the answer and do not output any other words. Question: \{input\}  Answer: & \small Answer the question based on the given passages. Only give me the answer and do not output any other words. If the question cannot be answered based on the information in the article, write “unanswerable”. The following are given passages. \{context\} Answer the question based on the given passages. Only give me the answer and do not output any other words. If the question cannot be answered based on the information in the article, write “unanswerable”. Question: \{input\} Answer: & \small Answer the question based on the given passages. The following are given passages. \{context\} Now, first provide your reasoning briefly in 2-3 sentences starting with 'Rationale:'. Then, answer the question starting with 'Answer:' based on the given passages. Question: \{input\} Rationale:\\ \hline
    MuSiQue & \small Answer the question based on the given passages. Only give me the answer and do not output any other words. The following are given passages. \{context\} Answer the question based on the given passages. Only give me the answer and do not output any other words. Question: \{input\}  Answer:& \small Answer the question based on the given passages. Only give me the answer and do not output any other words. If the question cannot be answered based on the information in the article, write “unanswerable”. The following are given passages. \{context\} Answer the question based on the given passages. Only give me the answer and do not output any other words. If the question cannot be answered based on the information in the article, write “unanswerable”. Question: \{input\} Answer: & \small Answer the question based on the given passages. The following are given passages. \{context\} Now, first provide your reasoning briefly in 2-3 sentences starting with 'Rationale:'. Then, answer the question starting with 'Answer:' based on the given passages. Question: \{input\} Rationale:\\ \hline
    QMSum & \small You are given a meeting transcript and a query containing a question or instruction. Answer the query in one or more sentences. Transcript:  \{context\} Now, answer the query based on the above meeting transcript in one or more sentences. Query: \{input\}  Answer: & \small You are given a meeting transcript and a query containing a question or instruction. Answer the query in one or more sentences. If the question cannot be answered based on the information in the article, write “unanswerable”. Transcript:  \{context\} Now, answer the query based on the above meeting transcript in one or more sentences. If the question cannot be answered based on the information in the article, write “unanswerable”. Query: \{input\}  Answer: & \small You are given a meeting transcript and a query containing a question or instruction. Answer the query in one or more sentences. Transcript:  \{context\} Now, answer the query based on the above meeting transcript in one or more sentences. Query: \{input\}  Answer:\\ \hline
  \end{tabular}}
  \caption{(Part 2 / 2) Prompts used in our experiments for LongBench datasets.}
  \label{tab:prompts-lb-2}
\end{table*}

\begin{table*}
  \centering
  \scalebox{0.8}{
  \begin{tabular}{l|p{4cm}|p{5cm}|p{4cm}}
    \hline
    \textbf{Dataset} & \textbf{LC, Vanilla / OP RAG} & \textbf{Self-Route: Stage I} & \textbf{FB-RAG: Stage II} \\ \hline
    EN.QA & \small Read the book and answer the question. Be very concise in your answer. Book: \{context\} Now, answer the question based on the book. Only give me the answer and do not output any other words. Question: \{input\} Answer: & \small Read the book and answer the question. Be very concise in your answer. If the question cannot be answered based on the information in the article, write “unanswerable”. Book: {context} Now, answer the question based on the book. Only give me the answer and do not output any other words. If the question cannot be answered based on the information in the article, write “unanswerable”. Question: \{input\} Answer: & \small Read the book and answer the question. Be very concise in your answer. Book: \{context\} Now, first provide your reasoning briefly in 2-3 sentences starting with 'Rationale:'. Then, answer the question starting with 'Answer:' as concisely as you can. Question: \{input\} Rationale:\\  \hline
    EN.MC & \small Read the book and answer the question. Book: \{context\} Now, answer the question based on the book. Only output the answer and do not output any other words. Question: \{input\}  \{all\_classes\} Answer: & \small Read the book and answer the question. If the question cannot be answered based on the information in the article, write “unanswerable”. Book: \{context\} Now, answer the question based on the book. Only output the answer and do not output any other words. If the question cannot be answered based on the information in the article, write “unanswerable”. Question: \{input\}  \{all\_classes\} Answer: & \small Read the book and answer the question. Book: \{context\} Now, first provide your reasoning briefly in 2-3 sentences starting with 'Rationale:'. Then, answer the question starting with 'Answer:' as concisely as you can. Question: \{input\}  \{all\_classes\} Rationale:\\ \hline
  \end{tabular}}
  \caption{Prompts used in our experiments for $\infty$Bench datasets.}
  \label{tab:prompts-ib}
\end{table*}

\section{Results}
\label{sec:appendix-results}

\begin{figure}[t]
\centering
  \includegraphics[width=0.9\columnwidth]{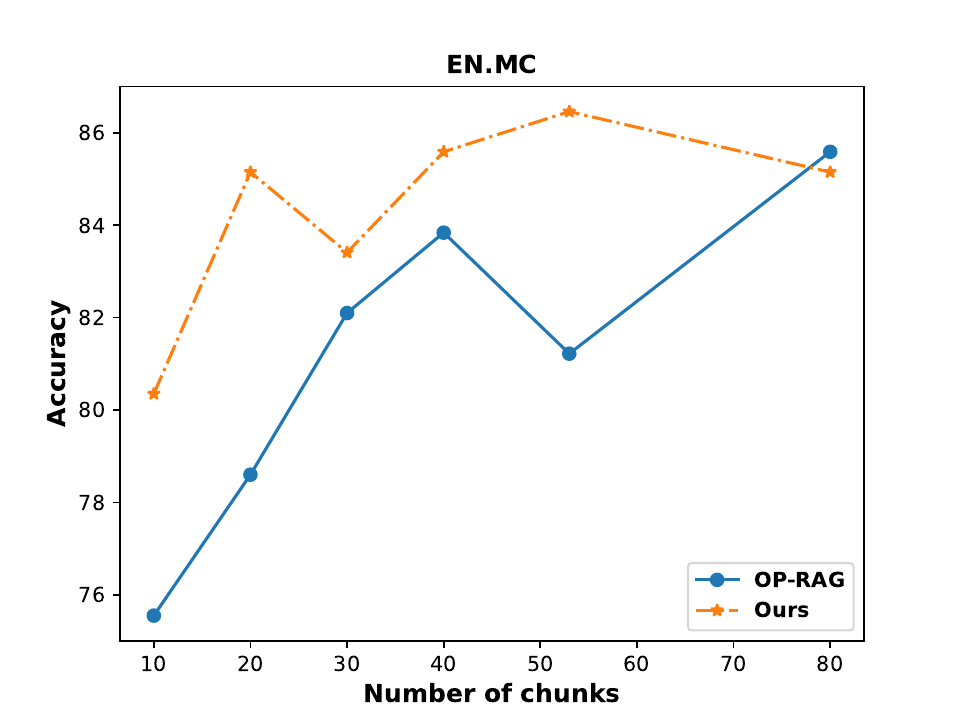}
  \caption{Performance comparison between our approach and OP RAG on \textbf{EN.MC} dataset. Y-Axis: The performance on the corresponding metric. X-Axis: The number of chunks used by both methods for final response generation. Across all data points, our approach uses an Llama3.1-8B-Instruct model for forward lookup in Stage 2 with $80$ context chunks as input and setting $\eta_F=1$ and $\eta_B=0$.}
  \label{fig:chunks-enmc}
\end{figure}

\begin{figure}[t]
 \centering
  \includegraphics[width=0.9\columnwidth]{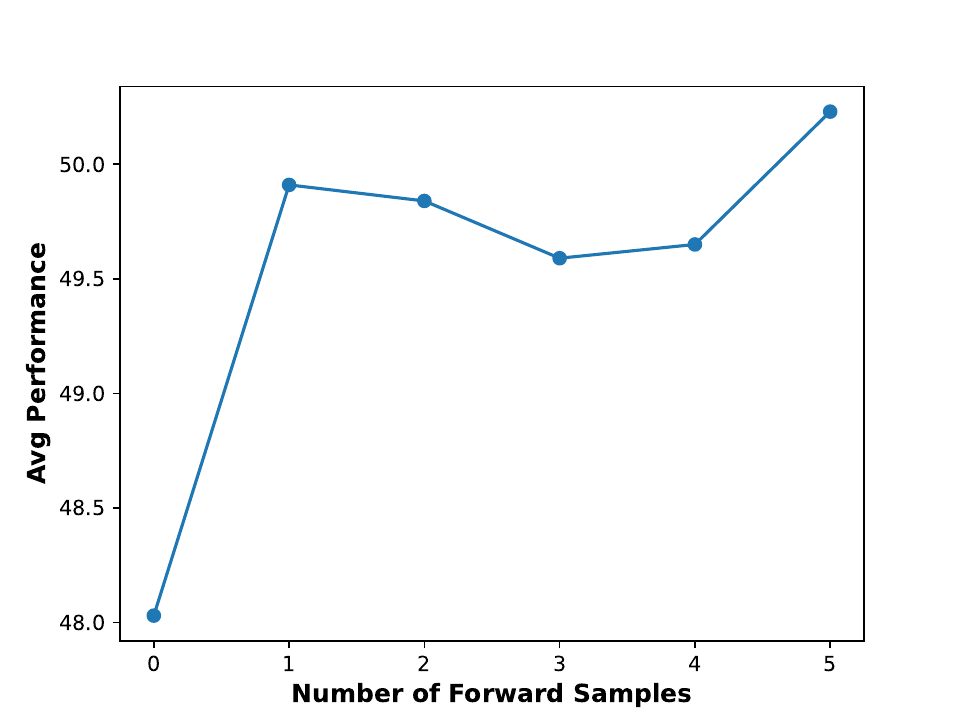}
  \caption{Studying the impact on the average performance of FB-RAG on LongBench datasets by varying the number of samples used in Stage II. Model used: Ours-FB (6k $\to$ 3k).}
  \label{fig:fwgens-longbench}
\end{figure}

\subsection{Retriever comparisons}
\label{sec:appendix-retriever-comparison}
We compared the performance of several off-the-shelf retrievers in our initial experiments, as presented in Table \ref{tab:appendix-retriever-longbench}. All methods use OP RAG at 3k context size. We find that BM25 performs reasonably well on average in comparison to numerous top-performing semantic retrievers. In addition, BM25 is a versatile approach without any underlying assumptions about the query, making it well-suited for our forward-looking approach in this paper. Hence, we fixed BM25 as the retriever for the rest of our experiments discussed in Section \ref{sec:results} in the main paper.

\subsection{FB-RAG Stage II Prompt comparisons}
\label{sec:appendix-prompt-comparisons}
We experimented with a few prompt variations for Stage II of FB-RAG. Table \ref{tab:appendix-prompt-vary} presents these comparisons on LongBench datasets. We observe that only using the answers sampled from the LLM shows improvements over other RAG baselines presented in the main paper, although this can be further improved slightly by using some form of reasoning along with the answer. This helps to handle scenarios where the answers are entity names or binary that contain little information for retrieving the most relevant context chunks.

\subsection{Varying the number of chunks used for final generation}
\label{sec:appendix-vary-chunks-enmc}

In Figure \ref{fig:chunks-enmc}, we compare the performance of our approach with OP-RAG on \textbf{EN.MC} dataset by varying the number of chunks used for final generation. We find that FB-RAG at $53$ chunks (16k context) beats the best performance of the baseline at $80$ chunks (24k context).

\subsection{Varying the number of samples used in Stage II of FB-RAG}
\label{sec:appendix-num-samples}

We present the plot for analysis in Figure \ref{fig:fwgens-longbench}. The X-axis denotes the number of samples used. The Y-axis denotes the average performance on LongBench datasets. The results are shown for the \textbf{Ours-FB (6k $\to$ 3k)} configuration. As evident from the figure, we find that the performance improves visibly with just one forward sample, while attaining the maximum at $5$ samples.

\subsection{In-Context Learning experiments}
\label{sec:appendix-combine-icl}

Section \ref{sec:discussion} discusses additional analysis where we combine the techniques developed in this work with the complementary methods proposed by \citet{wei2024instructrag}. We present the results from these experiments in Table \ref{tab:appendix-combine-icl}.

% Full result tables with all the results + any analysis tables that cannot fit, like comparing IR methods, comparing the rationale prompt.. Give BM25 implementation.

\begin{table*}
  \centering
  \scalebox{1.0}{
  \begin{tabular}{l|c|ccccccc}
    \hline
    \textbf{Method} & \textbf{Avg} & \textbf{Narr} &  \textbf{Qasp} & \textbf{Mult} & \textbf{Hotp} & \textbf{2Wiki} & \textbf{Musi} & \textbf{QMSum} \\
    \hline
    BM25 & 48.03 & 26.62 & 50.71 & 56.78 & 66.28 & 64.8 & 45.91 & 25.11\\
    M3Flag (1, 0, 0) & 48.3 & 29.4 & 50.36 & 55.99 & 63.76 & 66.47 & 47.87 & 24.23 \\
    M3Flag (1, 0.3, 0) & 48.58 & 29.79 & 50.14 & 55.86 & 64.83 & 66.78 & 48.33 & 24.36\\
    BGEFlag & 48.05 & 27.79 & 51.24 & 53.99 & 66.64 & 66.46 & 45.74 & 24.49 \\
    MPNet & 46.92 &  25.97 &  50.72 &  54.33 &  62.95 &  65.55 &  44.7 &  24.25\\ \hline
  \end{tabular}}
  \caption{Performance comparisons of off-the-shelf retrievers on LongBench datasets. All results are based on OP RAG at 3k context with Llama3.1-70B-Instruct model. We compared two weight configurations for M3Flag, taking recommendations from the authors to set the weights - refer to the original paper for details~\cite{chen-etal-2024-m3}.}
  \label{tab:appendix-retriever-longbench}
\end{table*}

\begin{table*}
  \centering
  \scalebox{1.0}{
  \begin{tabular}{l|c|ccccccc}
    \hline
    \textbf{Method} & \textbf{Avg} & \textbf{Narr} &  \textbf{Qasp} & \textbf{Mult} & \textbf{Hotp} & \textbf{2Wiki} & \textbf{Musi} & \textbf{QMSum} \\
    \hline
    Only answers & 50.09 & 30.63 & 52.11 & 56.17 & 66.16 & 68.97 & 51.49 & 25.07\\
    Thought process & 50.09 & 32.33 & 51.6 & 55.63 & 65.42 & 68.09 & 52.8 & 24.76\\
    Explanation & 50.33 & 30.83 & 51.84 & 55.88 & 66.92 & 68.62 & 53.67 & 24.54\\
    Reasoning & 50.23 & 33.22 & 50.99 & 55.99 & 66.29 & 67.42 & 53.13 & 24.56\\ \hline
  \end{tabular}}
  \caption{Performance comparisons of our approach on LongBench datasets by varying the prompt used for sampling in Stage II. Model Used: Ours-FB (6k $\to$ 3k). \textit{Thought process:} Generate the thought process before the final answer, \textit{Reasoning:} Generate a reasoning sequence before the final answer, \textit{Explanation:} Generate an explanation after generating the answer. While the performance improves over the baselines by only considering the final answers as samples, we find that using reasoning or explanation performs slightly better on average.}
  \label{tab:appendix-prompt-vary}
\end{table*}

\begin{table*}
\centering
\scalebox{0.8}{
\begin{tabular}{l|c|ccccccc}
\hline
\textbf{Method} & \textbf{Avg} & \textbf{Narr} &  \textbf{Qasp} & \textbf{Mult} & \textbf{Hotp} & \textbf{2Wiki} & \textbf{Musi} & \textbf{QMSum} \\
\hline
\multicolumn{9}{c}{Results copied from Table \ref{tab:results-longbench} in the main paper} \\
Vanilla & 44.19 & 25.01 & 49.31 & 53.41 & 60.91 & 58.84 & 37.32 & 24.51\\
Ours-F (6k $\to$ 1.5k) & \textbf{49.36} & \textbf{28.62} & \textbf{51.29} & \textbf{55.53} & \textbf{66.99} & 65.1 & \textbf{52.93} & \textbf{25.07} \\ \hline
\multicolumn{9}{c}{Augmented to Vanilla RAG} \\
Reason-then-Answer & 33.42 & 19.01 & 46.08 & 45.34 & 43.78 & 29.6 & 26.49 & 23.61 \\
Few-Shot Demo. w/ Instruction & 41.59 & 21.24 & 46.49 & 50.54 & 54.49 & 56.24 & 37.75 & 24.36 \\
INSTRUCTRAG-ICL & 42.95 & 23.06 & 47.55 & 53.74 & 54.8 & 61.47 & 35.93 & 24.08 \\ \hline
\multicolumn{9}{c}{Augmented to Ours-F (6k $\to$ 1.5k)} \\
 Reason-then-Answer & 37.71 & 27.74 & 48.48 & 45.33 & 46.61 & 37.06 & 35 & 23.75 \\
 Few-Shot Demo. w/ Instruction & 46.76 & 27.26 & 49.73 & 50.52 & 58.21 & 64.52 & 52.67 & 24.42 \\
INSTRUCTRAG-ICL & 48.79 & 28.2 & 48.99 & 55.52 & 63.7 & \textbf{69.42} & 51.51 & 24.22 \\ \hline
\end{tabular}}
\caption{Results on LongBench for 1.5k context length by combining \textbf{FB-RAG} with methods proposed in \citet{wei2024instructrag}. We use Rouge-L F1 for \textbf{QMSum}, and F1 score for others. (X $\to$ Y): Context size $X$ in Stage I and $Y$ in Stage II.}
\label{tab:appendix-combine-icl}
\end{table*}

\end{document}